\documentclass[letterpaper, 10 pt, conference]{ieeeconf}  % Comment this line out if you need a4paper

\IEEEoverridecommandlockouts                              % This command is only needed if
\usepackage{graphics} % for pdf, bitmapped graphics files
\usepackage{epsfig} % for postscript graphics files
\usepackage{mathptmx} % assumes new font selection scheme installed
\usepackage{times} % assumes new font selection scheme installed
\usepackage{amsmath} % assumes amsmath package installed
\usepackage{amssymb}  % assumes amsmath package installed
\usepackage{bm}

\DeclareMathOperator*{\argmin}{arg\,min}

\title{\LARGE \bf
Modular Safety-Critical Control of Legged Robots*
}

\author{Berk Tosun$^{1}$ and Evren Samur$^{2}$% <-this % stops a space
\thanks{*This work was supported by the Scientific and Technological Research Council of Turkey
(TUBITAK \#118E922)}% <-this % stops a space
\thanks{$^{1}$Graduate Program in Systems and Control Engineering, Bogazici University, Istanbul
        {\tt\small tb.tosunberk@gmail.com}}
\thanks{$^{2}$Department of Mechanical Engineering, Bogazici
        University, Istanbul {\tt\small evren.samur@boun.edu.tr}}%
}

\begin{document}

\maketitle
\thispagestyle{empty}
\pagestyle{empty}

%%%%%%%%%%%%%%%%%%%%%%%%%%%%%%%%%%%%%%%%%%%%%%%%%%%%%%%%%%%%%%%%%%%%%%%%%%%%%%%%
\begin{abstract}

Safety concerns during the operation of legged robots must be addressed to enable
their widespread use. Machine learning-based control methods that use model-based constraints provide promising means to
improve robot safety.
This study
presents a modular safety
filter to improve the safety of a legged robot, i.e., reduce the chance of a fall. The prerequisite is the
availability of a robot that is capable of locomotion, i.e., a nominal controller exists. During locomotion, terrain properties around the robot are
estimated through machine learning
which uses a minimal set of
proprioceptive signals. A novel deep-learning model utilizing an efficient transformer architecture is used for the terrain estimation.
A quadratic program combines the terrain estimations with inverse dynamics and a novel exponential control barrier function constraint
to filter and certify nominal control signals.
The result is an optimal controller that acts as a filter. The filtered control signal allows safe locomotion of the robot.
The resulting approach is generalizable, and could be transferred with low effort to any other legged system.

\end{abstract}

%%%%%%%%%%%%%%%%%%%%%%%%%%%%%%%%%%%%%%%%%%%%%%%%%%%%%%%%%%%%%%%%%%%%%%%%%%%%%%%%
\section{Introduction}

As in nature, legs enable robots to travel in challenging environments such as rough terrain, to climb stairs,
and to reach tight spots. The next step for legged robots
is integrating them further into daily lives with an emphasis on safety.

In the scope of this work, we consider the safety of a legged robot as maintaining its ability to locomote;
it can be also stated simply but roughly as not falling.
It must be noted that
there is no concrete definition of safe locomotion as there is no
agreed way of quantifying safety~\cite{Grizzle2021Models,Zaytsev2018}.

\begin{figure}[thbp]
  \centering
    \includegraphics[width=0.48\textwidth, keepaspectratio]{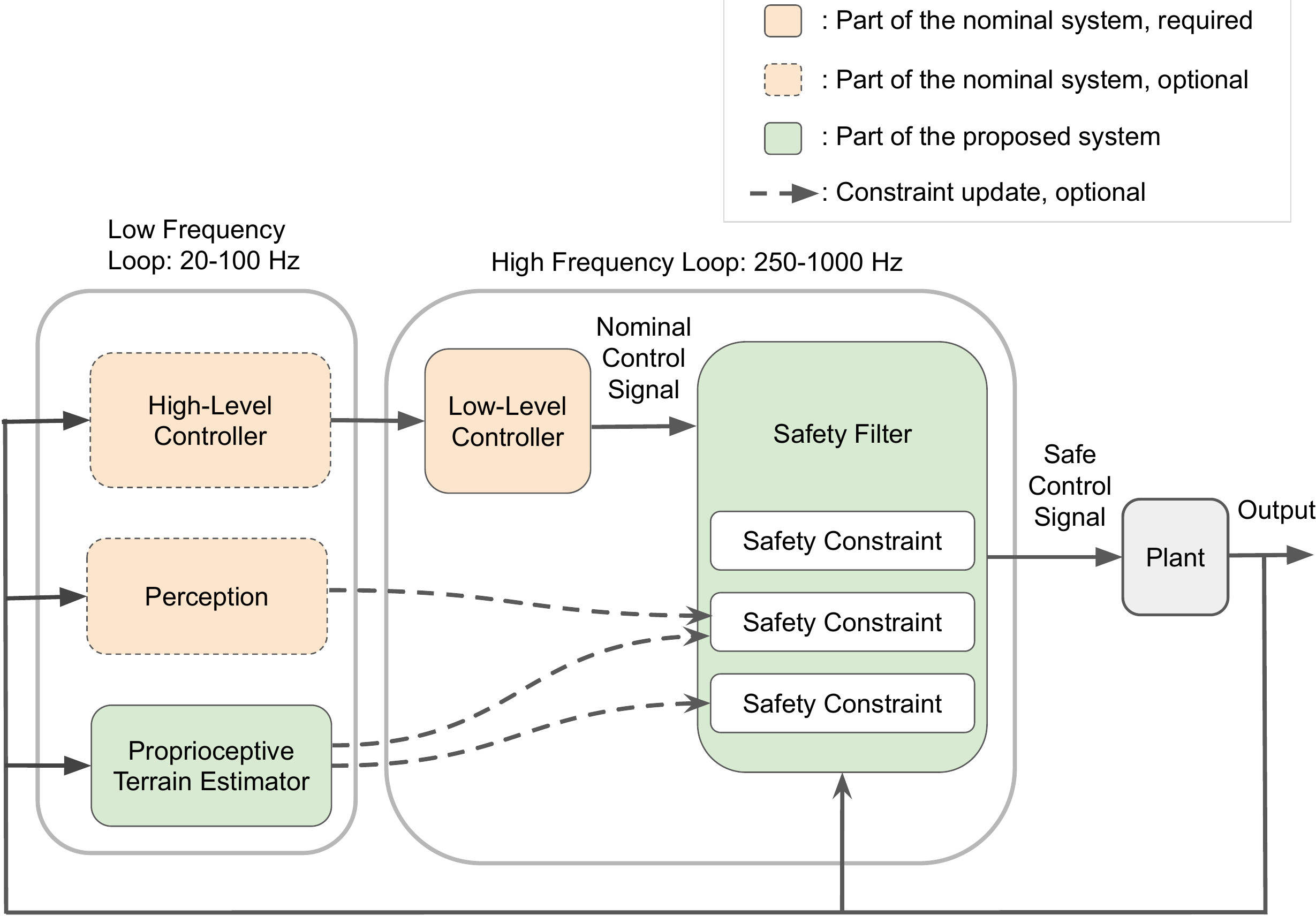}
    \caption{Block diagram of a two-level closed-loop control system, as commonly found in robotics.
    Our proposed system is shown as a modular safety filter that is added with minimal modification to an existing architecture.}
  \label{fig:modular_diagram}
\end{figure}

\subsection{Related Work}

In current literature, the stability of legged robots is
considered for flat ground.
However, this assumption is often
violated; the point of having legs is to clear rough terrain.
As a solution, some recent
work focused on estimating the effect of the terrain~\cite{Carpentier2018}.
Others~\cite{Siekmann2021BlindStair,Miki2022perfect-rl} have shown only proprioceptive, i.e., blind, measurements
combined with deep learning are effective in considering terrain effects. However,
there has been no work on a dedicated proprioceptive terrain estimator. In this study, we propose a method such that
the robot gains access to the terrain properties with no additional sensors.
With the proposed approach, only the sensors required to control a robot
should suffice; thus reducing the cost and complexity due to sensors like
cameras and lidars.

Another recent development is
the introduction of the control barrier function~(CBF) framework to guarantee the safe operation of dynamic systems
such as legged robots~\cite{Ames2019}.
Our work uses the CBF framework to generate constraints such that rough terrain can be safely navigated. The terrain can be estimated
by the previously described blind terrain estimator, and fed into the constraints to generate an adaptive safety filter.
The resulting system will function as a filter, allowing it to be integrated easily as shown in Figure~\ref{fig:modular_diagram}.

Some of the most impressive, high-performance controllers
are obtained with reinforcement-learning methods~\cite{Miki2022perfect-rl}. However, learning-based controllers are based on
simple models or no model at all, i.e., model-free. Combining such controllers with model-based safety
constraints would be an effective way to address safety concerns. Thus, our proposed safety filter
would be especially valuable in the testing and eventual deployment of learning-based systems.

\subsection{Contributions}

In this work, the following three contributions are made:
\begin{itemize}
    \item An intuitive, high-level optimal-control-based safety filter framework is implemented with
    direct support for the CBF framework.
    The source code is publically available~\cite{source-code}.
    \item The transformer architecture from deep learning is adapted
    to model terrain friction using only proprioceptive measurements.
    \item A novel exponential control barrier function~(ECBF) is formulated and implemented for ground clearance of the swing foot of legged robots.
\end{itemize}

\begin{figure}[thpb]
    \centering
    \includegraphics[scale=0.35]{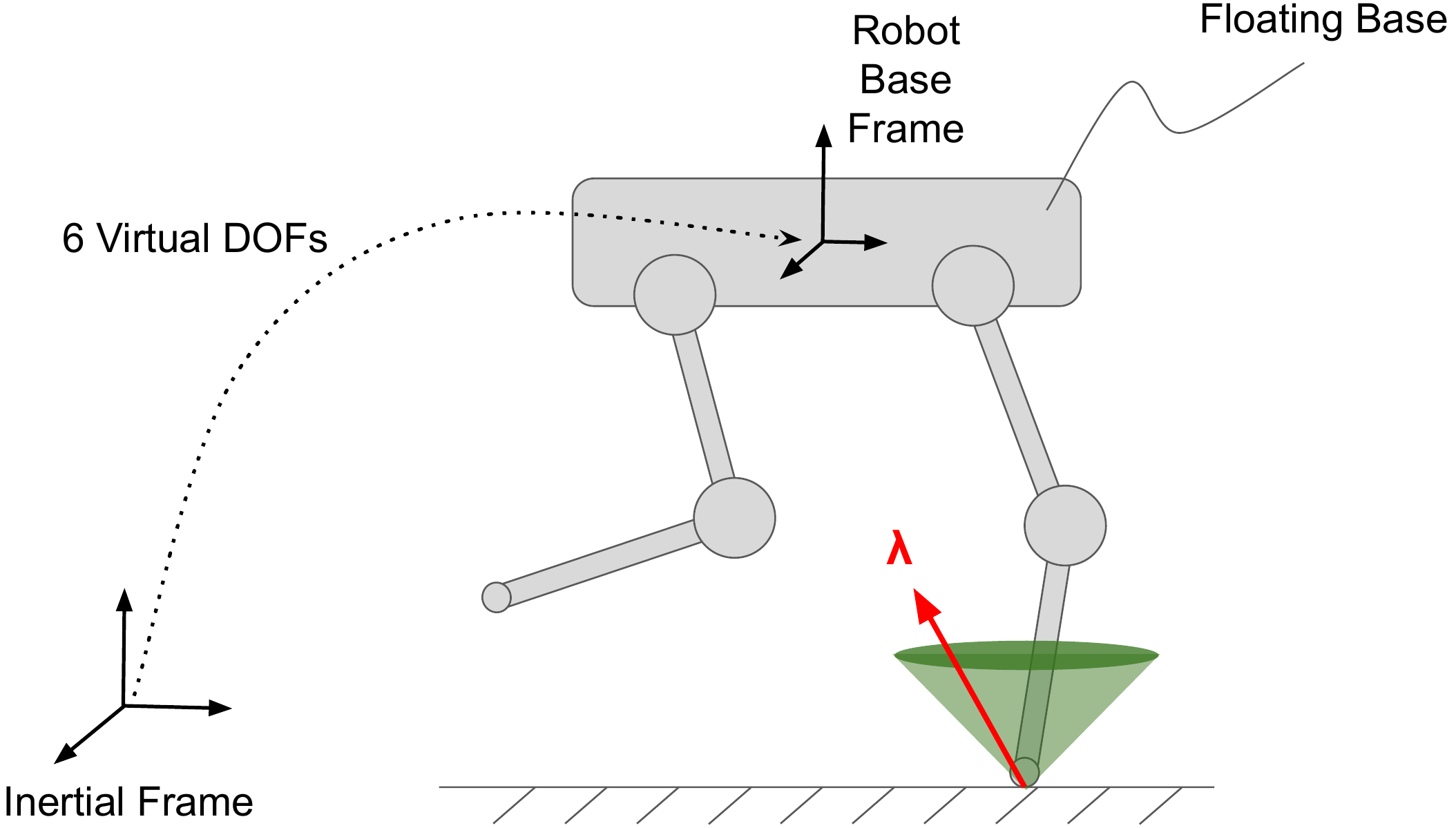}
    \caption{Floating base robot with one point foot in contact.
    The foot is in stance mode, as the reaction force $\lambda$ lies inside the friction cone}
    \label{fig:flight_coords}
\end{figure}

\section{Background}

\subsection{Dynamics of a Legged Robot}

A legged robot, as shown in Figure~\ref{fig:flight_coords}, is an underactuated control system~\cite{underactuated}.
Define $n_{va}$ as the number of actuated joints and $n_{vu}$ as the number of unactuated joints.
The total DoF is defined as $n_v = n_{va} + n_{vu}$. Underactuation implies $n_{va} < n_v $;
robot control must be achieved through contact.

In the scope of robot dynamics~\cite{Featherstone2008Dynamics,Wieber2006dynamics}, contact is commonly modeled as rigid point contacts.
For legged robots, which make and break
contact in various combinations, a popular choice is to use excess coordinates. In this case,
a link is selected as the floating base; then the floating base frame and an inertial frame are
connected through 6 virtual, unactuated DoFs~(degrees-of-freedom), $\bm q_{\bm u}$. The resulting equation allows us to represent the dynamics concisely:
\begin{equation}
    \bm M(\bm q)\dot{\bm v} + \bm H(\bm q, \bm v) = \bm B \bm u + \bm J_{\bm c}(\bm q)^T \bm \lambda,
\label{eq-eom-flight}
\end{equation}
where
\begin{itemize}
  \item $\bm q \in \mathbb{R}^{n_{q_u} + n_{va}}$ is the robot configuration vector which includes both actuated and unactuated DoFs;
  $\bm q = (\bm q_{\bm u}, \bm q_{\bm a})$.
%   $n_{\bm q_{\bm u}}$ is the number of parameters used to represent
%   floating base configuration. Inline with the literature, we use quaternions to represent base orientation. Thus $n_{\bm q_{\bm u}} = 7$; 3 to represent
%   translation, 4 for orientation.
\item $\bm v \in \mathbb{R}^{6 + n_{va}}$ is the robot velocity, $\bm v = (\bm v_{\bm u}, \bm v_{\bm a})$.
% Note $\dot{\bm q}\neq \bm v$ unless Euler angles are used to represent base orientation.
% $\bm q_u \in SE(3)$ includes rotations that can be represented in different ways; regardless of
% the configuration of $q_u$, the velocity of the floating base is represented by six parameters.
\item $\dot{\bm v} \in \mathbb{R}^{6 + n_{va}}$ is the robot accelerations.
\item $\bm u \in \mathbb{R}^{n_{va}}$ is the control inputs, i.e., joint torques.
  \item $\bm M(\bm q) \in \mathbb{R}^{n_v \times n_v}$ is the mass matrix.
  \item $\bm H(\bm q, \bm v)  \in \mathbb{R}^{n_v}$ is the nonlinear effects: centrifugal, Coriolis, and gravity terms.
  \item $\bm B \in \mathbb{R}^{n_v \times n_{va}}$ is the selection matrix, $\bm B = [\bm 0_{n_{va} \times n_{vu}} \ \bm I_{n_{va}}]^T$.
  \item $\bm J_{\bm c}(\bm q)^T \in \mathbb{R}^{n_v \times 3n_c}$ is the contact jacobian transposed. Define $n_c$ as the number of contact
  points. For each contact point, there will be a 3D constraint force.
  \item $\bm \lambda \in \mathbb{R}^{3n_c}$ is the vector of constraint forces. It is updated whenever there is a change in the contact states.
  \item $\bm J_{\bm c}(\bm q)^T\lambda$ is the vector of forces applied on the robot DoFs due to contact. It is the only
  term to control $\bm q_{\bm u}$.
\end{itemize}

\begin{figure}[thbp]
  \centering
    \includegraphics[width=0.48\textwidth, keepaspectratio]{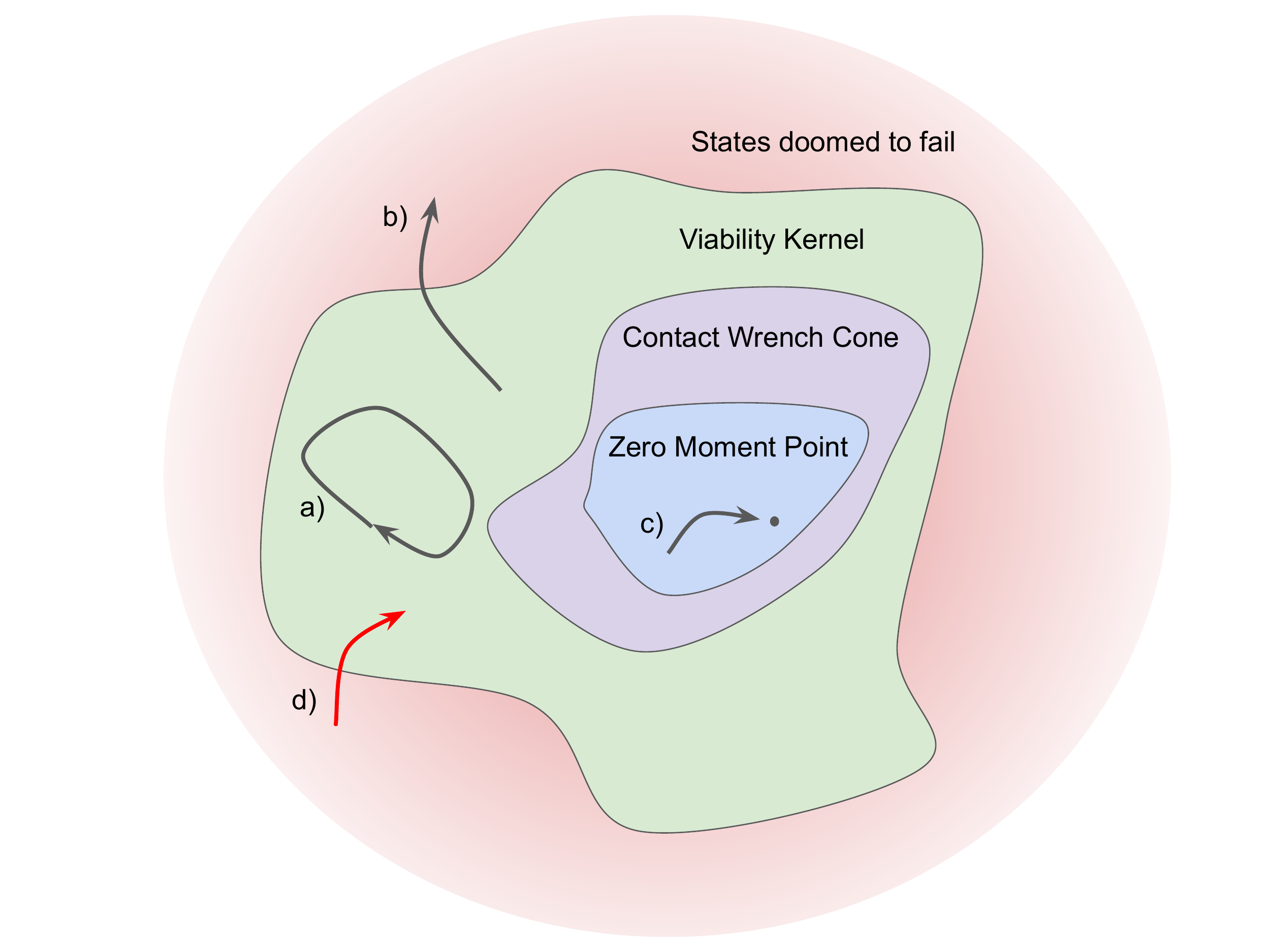}
    \caption{Evolution of some sample states and their relation with the viability kernel:
    a)~Limit cycle, stable walking; b)~Robot leaving the viability kernel, e.g., it tips over;
    c)~Robot takes a single step while remaining statically stable for the whole duration;
    d)~Impossible state transition.}
  \label{fig:viability}
\end{figure}

\subsection{Quantifying Safety: Viability}

For traditional control systems, stability is analyzed by established measures such as eigenvalues or phase margins.
Legged robots require special attention due to the complexity of quantifying the robot's safety.
Safe locomotion of legged robots is seen as a problem of viability instead of Lyapunov
stability~\cite{Wieber2008,Zaytsev2018}.

The viability kernel is defined as the set of all states where the system remains safe,
i.e., it can keep operating.
% For legged robots, all states where the robot can keep its ability to locomote are
% included in the
% system's viability kernel.
By definition, any
state that is out of the viability kernel cannot return to the viability kernel, and thus it eventually ends up in a failed state.
Figure~\ref{fig:viability} illustrates this via examples.

% The established safety criteria have their uses. However, their effect on the resulting closed-loop control system must
% be considered thoroughly. For example, ZMP requires the robot to always remain statically stable. Such control results in
% unnatural, slow, and highly energy-inefficient locomotion. CWC improves on ZMP and
% allows the controller to use a wider set of the viability kernel.
Recent literature~\cite{Miki2022perfect-rl,Levine2021Lessons} shows that learning-based robot controllers
overperform the classical controllers synthesized by the established safety criteria~\cite{Vukobratovi2004zmp,Hirukawa2006CWC}.
Learning-based ones do not necessarily consider any such safety criteria.

% \begin{figure}[thbp]
%   \centering
%     \includegraphics[width=0.48\textwidth, keepaspectratio]{figures/safety_filter.pdf}
%     \caption{A safety filter inserted into an existing control system to guarantee safety.}
%   \label{fig:safety_filter}
% \end{figure}

\subsection{Control Barrier Functions}

The CBF framework~\cite{Ames2019} lends itself to the synthesis of controllers which possess
set forward invariance.
As a model-based approach, it enables theoretical guarantees by checking and restricting derivatives.
Define the system states as $\bm{x} \in D$,
where $D$ is the entire state space of the system.
Then a CBF, $h: D \rightarrow \mathbb{R}$, is defined so that returned scalar value must always be positive for safe states,
i.e., safe if $h \geq 0$. The framework allows the construction of linear inequality constraints that can be used in optimal controllers.

\begin{figure}[thbp]
  \centering
    \includegraphics[scale=0.4]{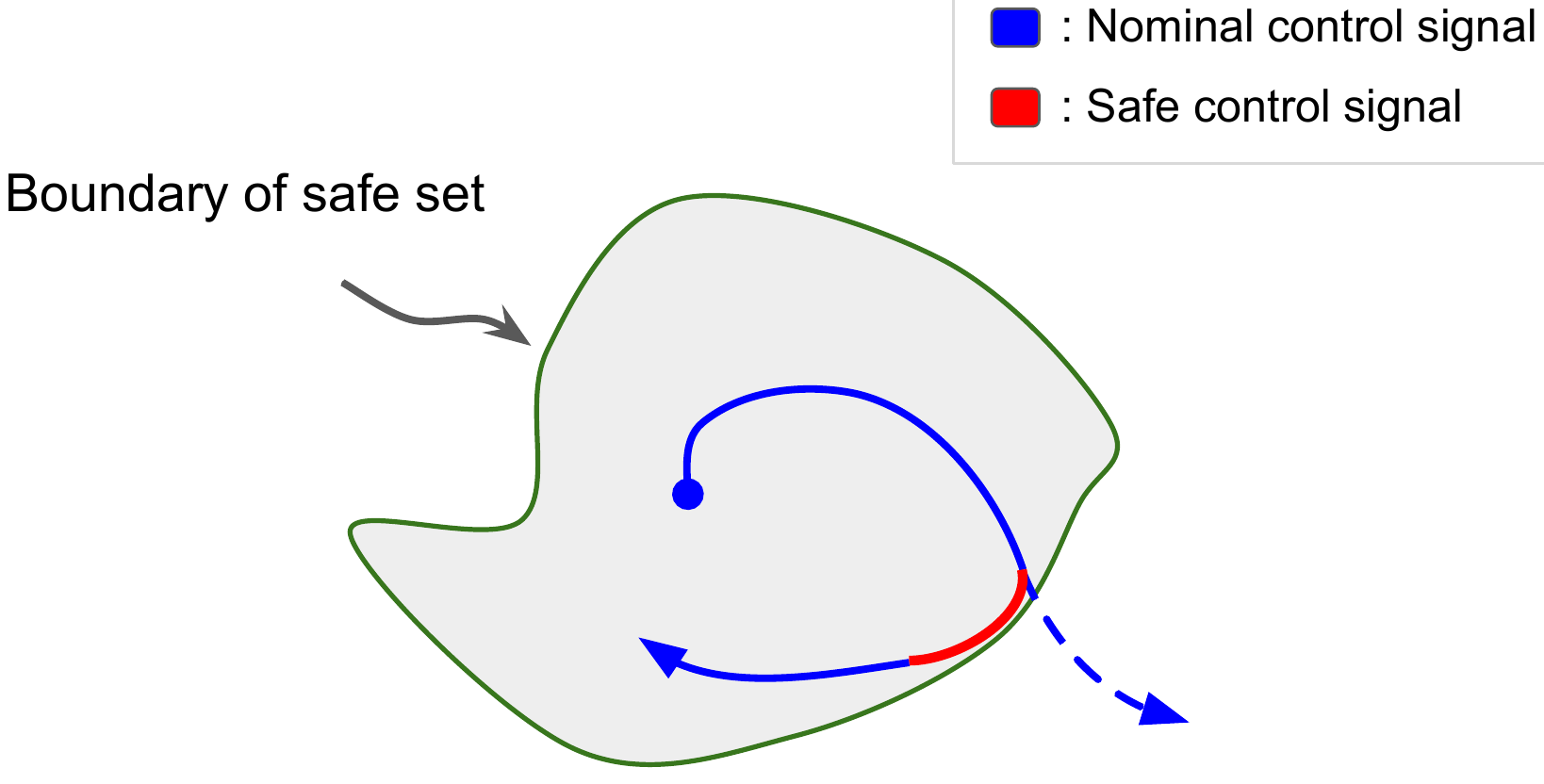}
    \caption{Evolution of the states of a dynamic system displaying set forward invariance with the help of a safety filter.
    The safety filter plays an active role near the boundary to keep the system in a safe set.}
  \label{fig:safety_filter_evolution}
\end{figure}

As shown in Figure~\ref{fig:modular_diagram},
% As shown in Figure~\ref{fig:safety_filter},
the CBF framework is commonly used as a safety filter that can be inserted into
an existing control system after the
nominal controller. In a quadratic program~(QP), we can include linear inequality constraints allowing us to formulate a CBF-QP safety filter.
The safety filter is minimally intrusive, i.e., it will only interfere with the nominal control signal near the boundary of the
safe set to make sure the system remains in the safe states, which is illustrated in Figure~\ref{fig:safety_filter_evolution}.

\subsection{Terrain Estimation with Spatial-Temporal Data}
\label{subsection:model-terrain}

To model the motion of agents, e.g., skeleton-based action recognition, it is common to use joint measurements, such as 3D coordinates, to
construct a graph~\cite{Yan2018}. The quantities for each joint at each time step are represented as a node in the graph.
Since the nodes are distributed over both spatial and temporal dimensions, connecting the nodes results in
a spatial-temporal model. Such graphs can be processed by machine learning models.

Transformer architecture has retained its popularity in deep
learning since its introduction by~\cite{Vaswani2017}.
The self-attention of the transformer gives it
its expressive power, but it is costly to compute. Its limitation of modeling long sequences can be overcome with
efficient transformers~\cite{Tay2020,Wang2020}.

% The regular transformer with the self-attention mechanism is roughly equivalent to a graph network working on a fully-connected graph~\cite{joshi2020transformers}.
% Extending the discussion to the efficient transformer architectures, we can think of sparse graphs instead of fully-connected ones.

Previous work has shown that it is possible to estimate
terrain properties by only using proprioceptive measurements~\cite{Lee2020}. Their analysis shows
that the model can accurately reconstruct the terrain. With terrain properties,
two important ones affecting the robot dynamics are meant: friction coefficient and a rough estimate of the elevation
map of the terrain, i.e., macroscopic surface roughness.

\section{Modular Safety-Critical Control}

\subsection{Inverse Dynamics Formulation with CBF Constraints}
\label{subsection:id-cbf-qp}

To implement a safety filter, one can use the CBF-QP formulation~\cite{Ames2019}. In that case, we must embed the dynamics into the barrier inequality constraints.
This requires two
undesirable operations: inversion of the mass matrix and solving for~$\lambda$~\cite{Reher2020}. Finding $ \lambda $ is challenging; it
requires the assumption that kinematic contact constraints are satisfied. Inversion of the mass matrix tends to
get numerically stiff~\cite{Featherstone2004}. Thus, it is preferable to have an inverse dynamics formulation without these
issues. Following~\cite{Grandia2021}, we achieve the following formulation, which will be called ID-CBF-QP:
\begin{align}
  &\argmin_{\bm X = (\dot{\bm v}, \bm u, \bm \lambda)} \frac{1}{2} || \bm u - \bm {u_{nominal}} ||^2_2
  \label{eq-qrd-objf}
  \\
  &s.t. \hspace{.2in} \bm M(\bm q)\dot{\bm v} + \bm H(\bm q, \bm v) = \bm B \bm u + \bm{J_c^T} \bm \lambda
  \label{eq-qrd-cstr-eom}
  \\
  &\hspace{.4in} \bm J_{\bm c}(\bm q)\dot{\bm v} + \dot{\bm J}_{\bm c}(\bm q) \bm v= 0
  \label{eq-qrd-cstr-contact}
  \\
  &\hspace{.4in} \lambda_z^{\{i\}} > 0
  \label{eq-qrd-cstr-unilateral}
  \\
  &\hspace{.4in} \tilde{\mu} \lambda_z^{\{i\}} \geq | \lambda_x^{\{i\}} |
  \label{eq-qrd-cstr-fcone-x}
  \\
  &\hspace{.4in} \tilde{\mu} \lambda_z^{\{i\}} \geq | \lambda_y^{\{i\}} |
  \label{eq-qrd-cstr-fcone-y}
  \\
  &\hspace{.4in} \dot{h}(\bm q, \bm v, \bm u) \geq -\alpha(h(\bm q, \bm v))
  \label{eq-qrd-cstr-cbf}
  \\
  &\hspace{.4in} -\bm \tau_{\bm {max}} \leq \bm u \leq \bm \tau_{\bm {max}},
  \label{eq-qrd-cstr-torque}
\end{align}
where
\begin{figure}[thbp]
  \centering
    \includegraphics[width=0.48\textwidth, keepaspectratio]{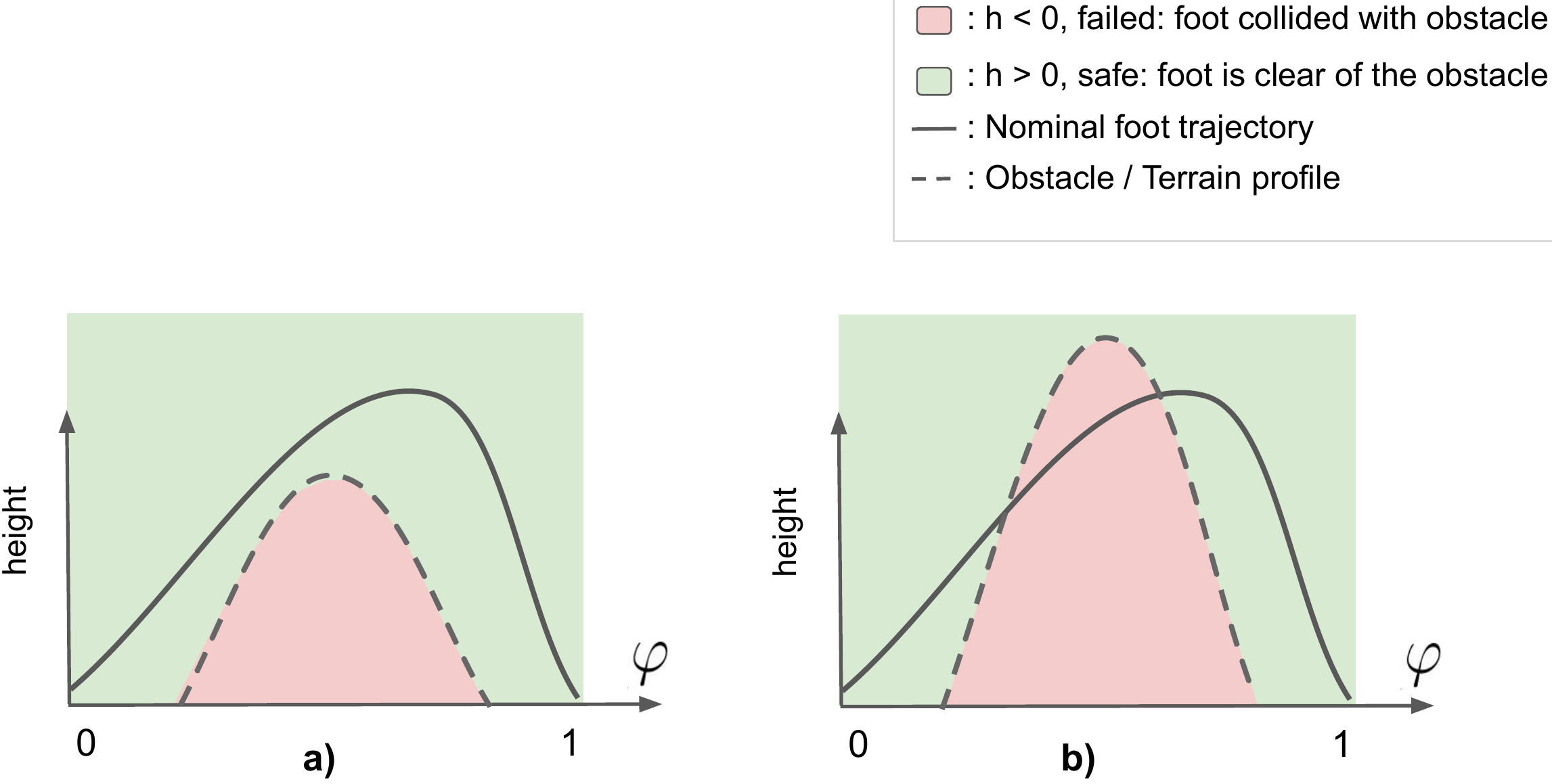}
    \caption{Two independent, sample cases of the ground-clearance ECBF for an arbitrary obstacle for a single foot.
    a) Obstacle is already cleared by the nominal trajectory, no interference is required.
    b) Obstacle must be cleared by adjusting the nominal trajectory, the ECBF can do it in an optimal sense.}
  \label{fig:ground-clearance-cstr}
\end{figure}
\begin{itemize}
  \item Equation~(\ref{eq-qrd-objf}) is the objective function. Three different decision variables, concatenated into vector $\bm X$,
  will be optimized. The decision variables are:
  \begin{enumerate}
    \item $ \dot{\bm v} \in \mathbb{R}^{n_{v}} $, accelerations for all degrees of freedom,
    \item $\bm u \in \mathbb{R}^{n_{va}} $, torques for actuated joints,
    \item $\bm \lambda \in \mathbb{R}^{3n_c} $, concatenated ground reaction forces from each contact point.
  \end{enumerate}
%   Note, $\dot{\bm v}$ can be included in the objective; for example, it can replace $\bm u - \bm {u_{nominal}}$ with
%   $\dot{\bm v} - \dot{\bm v}_{\bm {desired}}$. Or both $\bm u$ and $\bm v$ can be used with weights. This flexibility is one big advantage
%   of optimization-based control.
  \item Constrained equations of motion are included as constraints:
  \begin{enumerate}
    \item Equation~(\ref{eq-qrd-cstr-eom}) is the generalized equation of motion.
    \item Equation~(\ref{eq-qrd-cstr-contact}) is the kinematic constraint to keep contact points stationary.
  \end{enumerate}
  \item Stance feet must maintain their contact mode; it can be done by including friction cone constraints:
  \begin{enumerate}
    \item Equation~(\ref{eq-qrd-cstr-unilateral}) is the unilateral contact constraint to avoid loss of contact for each stance foot,
    foot ${\{i}\}$.
    \item Equation~(\ref{eq-qrd-cstr-fcone-x}) and (\ref{eq-qrd-cstr-fcone-y}) are the linearized friction cone
    constraints to avoid slipping for each stance foot, foot $y$.
    We use a square pyramid to get an inner approximation; it requires the use of $\tilde{\mu} = \mu/\sqrt{2}$
    as the effective friction coefficient.
  \end{enumerate}
  \item Equation~(\ref{eq-qrd-cstr-cbf}) is the control barrier constraint.
  \item Equation~(\ref{eq-qrd-cstr-torque}) is the joint torque limits.
\end{itemize}

\subsection{Ground Clearance Enforced by an ECBF}
\label{section:ECBF}

As an inequality constraint, we can incorporate CBFs into the optimization problem, as shown in (\ref{eq-qrd-cstr-cbf}).
There can be multiple CBF constraints as long as the QP remains feasible. In this instance, we define a novel CBF
constraint to keep the feet from hitting the ground by including the terrain profile.

We apply an ECBF to enforce ground-clearance of the robot's end-effectors, i.e., its swing feet; so that its feet do not
hit obstacles or adepts to uneven terrain while walking. Figure~\ref{fig:ground-clearance-cstr} describes the constraint by
illustrating two sample cases. We start the formulation with a position-based CBF:
\begin{equation}
    h(\bm q, \varphi) =\ ^{W}\bm p^{\{j\}}_z(\bm q) - z^{\{j\}}(\varphi),
    \label{eq-cbf-groundClearance}
\end{equation}
where $\varphi \in [0, 1]$ is the phase variable, denoting the progress of the swing phase;
$ ^{W}\bm p^{\{j\}}_z(\bm q): \mathbb{R}^{n_{v}} \rightarrow \mathbb{R}$ is the z-axis position (height) of the ${\{j}\}$th swing foot in the world frame;
$z^{\{j\}}(\varphi):\mathbb{R} \rightarrow \mathbb{R}$ maps the phase variable to desired ground-clearance height for the ${\{j}\}$th swing foot in the world frame.
${z(\varphi)}$ can be thought
as the obstacle heights over the gait period.
% Note, both $\bm q$ and $\varphi$ are functions of time, $t$.

\begin{figure}[thbp]
  \centering
    \includegraphics[width=0.48\textwidth, keepaspectratio]{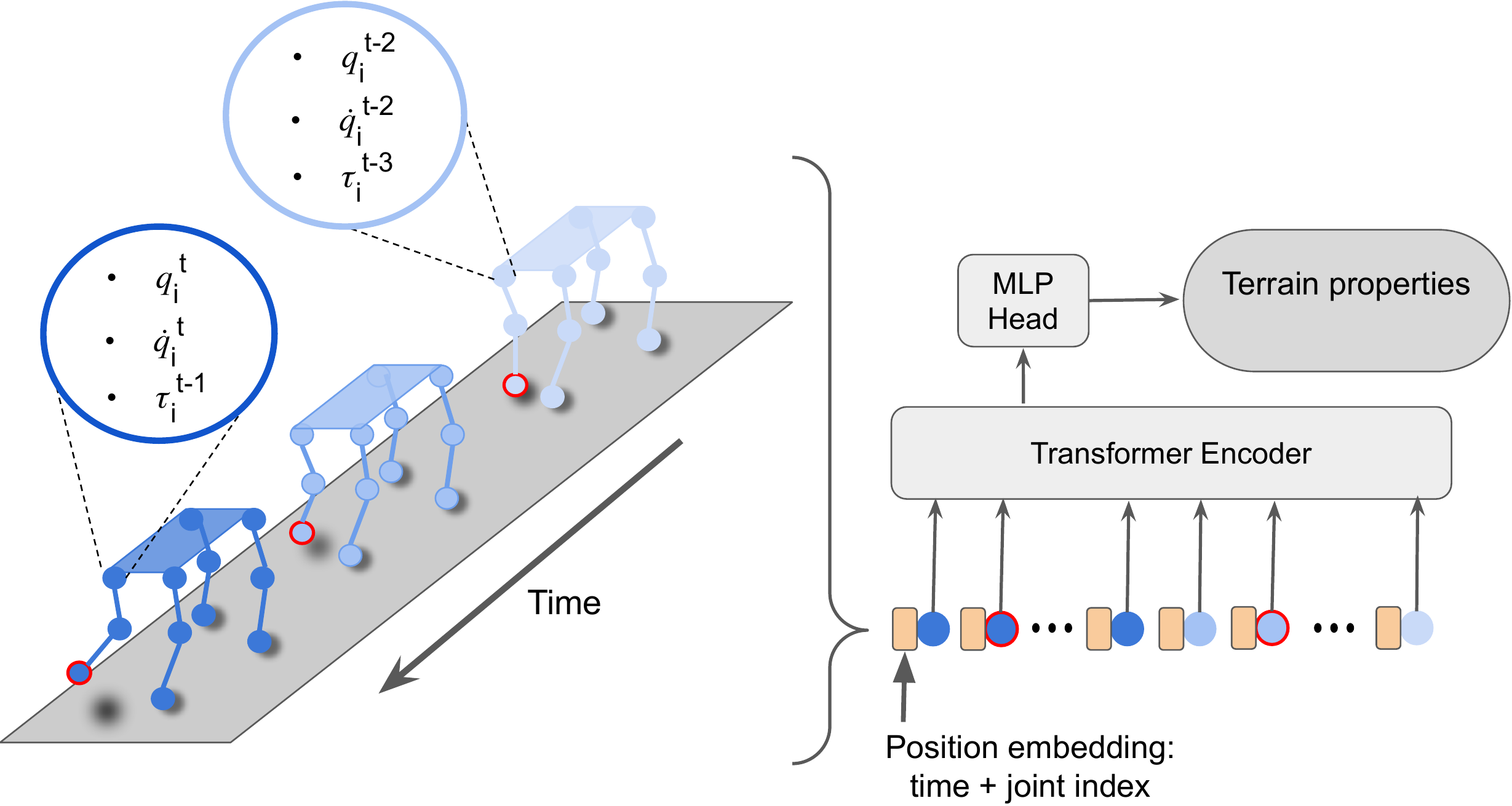}
    \caption{The robot moves one of its feet, shown with red, over time.
    Records for each joint at each time step are gathered and fed into the neural network. The network
    predicts terrain properties.}
  \label{fig:transformer_arch}
\end{figure}

To use the ECBF approach, we need $\varphi$ and its derivatives. In this work, we define the phase variable, $\varphi$ as follows:
\begin{equation}
    \varphi(t) =  \frac{t \mod period_{gait}}{period_{gait}},
\end{equation}
where $t$ is time. Its time derivates are given by:
\begin{align}
    &\dot{\varphi} = \frac{1}{period_{gait}},
    \\
    &\ddot{\varphi} = 0.
\end{align}
Since the ground-clearance CBF, $h$, is a position-based CBF, it cannot be directly used~\cite{Nguyen2016stepping}.
We encounter a control barrier function for a second relative degree safety constraint.
We solve this problem by using the generalized formulation, ECBF. To employ the ECBF approach we define an
auxiliary function,~$h_e$:
\begin{equation}
  h_e(\bm q, \bm v, \varphi) = \dot{h}(\bm q, \bm v, \varphi) + \alpha_1 h(\bm q, \varphi).
\end{equation}
Then, the ECBF constraint is:
\begin{equation}
  \alpha_2 h_e(\bm q, \bm v, \varphi) + \dot{h}_e(\bm q, \bm v, \dot{\bm v}, \varphi) \geq 0.
  \label{eq-ecbf-clearance}
\end{equation}
Taking the derivatives and substituting, we get:
\begin{equation}
    \begin{split}
  \alpha_2 h_e(\bm q, \bm v, \varphi) +\ ^{W}\ddot{\bm p}^{\{j\}}_z(\bm q) \\ - \ddot{z}^{\{j\}}(\varphi) \dot{\varphi}
  + \alpha_1 (^{W}\dot{\bm p}^{\{j\}}_z(\bm q) - \dot{z}(\varphi)^{\{j\}}\dot{\varphi}) \geq 0.
    \end{split}
\end{equation}
With the help of $\bm J_{\bm {flight}}$, Jacobian of the flight feet, we recover one of the optimization variables; $\dot{\bm v}$:
\begin{equation}
    \begin{split}
  \alpha_2 h_e(\bm q, \bm v, \varphi) +\ ^{W}(\dot{\bm J}^{\{\bm j\}}_{\bm {flight}}(\bm q)\bm v + \bm J^{\{\bm j\}}_{\bm {flight}}(\bm q)\dot{\bm v})_z \\ - \ddot{z}^{\{j\}}(\varphi)\dot{\varphi}
  +\ \alpha_1 (^{W}\dot{\bm p}^{\{j\}}_z(\bm q) - \dot{z}^{\{j\}}(\varphi)\dot{\varphi}) \geq 0.
  \label{eq-ecbf-clearance-final}
    \end{split}
\end{equation}
Note, the ECBF constraint, (\ref{eq-ecbf-clearance}), does not directly include the control signal, $\bm u$.
However, the CBF framework requires $\bm u$ in the final constraint. Equation~(\ref{eq-ecbf-clearance}) can serve as a valid CBF because,
$\dot{\bm v}$ appears in affine relation to $\bm u$ in (\ref{eq-qrd-cstr-eom})~\cite{Grandia2021}.

\begin{table}[thbp]
\vskip\baselineskip
\caption{Preprocessed input data for each batch.}
\begin{center}
\begin{tabular}{|c||c|} \hline
\textbf{Parameter} & \textbf{Value}\\\hline
  dt & 0.03 s\\\hline
  Number of timesteps & 40\\\hline
  Total time & 1.2 s\\\hline
  Number of environments (robots) & 4096\\\hline
  Number of features per timestep & 36\\\hline
\end{tabular}
\label{table:isaac-collect-subsample}
\end{center}
\end{table}

We use readily available, highly optimized QP solvers; as long as the inequality constraint is written in the standard
form, we can use any off-the-shelf solver:
\begin{equation}
  \bar{\bm G} \bm X \leq \bar{\bm h},
  \label{eq-standard-ineq}
\end{equation}
where $\bar{\bm G}$ is the linear inequality matrix, $\bar{\bm h}$ is the linear inequality vector.
% note $\bar{\bm G}$, $\bar{\bm h}$ are not related to $\bm G$ and $h$, we use bar to avoid any confusion.
Reorganizing (\ref{eq-ecbf-clearance-final})
to match the format of (\ref{eq-standard-ineq}), we get the inequality constraint in the standard form:
\begin{equation}
\small
    \begin{split}
    \begin{bmatrix}
        -\bm J^{\{\bm j\}}_{\bm {flight}}(\bm q) & \mathbf{0}_{1 \times n_{va}} & \mathbf{0}_{1 \times 3n_c}\\
        -\bm J^{\{\bm {j+1}\}}_{\bm {flight}}(\bm q) & \mathbf{0}_{1 \times n_{va}} & \mathbf{0}_{1 \times 3n_c}\\
        \vdots & \vdots & \vdots
    \end{bmatrix}
    \begin{bmatrix}
        \dot{\bm v} \\
        \bm u \\
        \bm \lambda
    \end{bmatrix} \leq \\
    \begin{bmatrix}
        \alpha_2 h_e(\bm q, \bm v, \varphi) +\ ^{W}(\dot{\bm J}^{\{\bm j\}}_{\bm {flight}}(\bm q)\bm v )_z - \ddot{z}^{\{j\}}(\varphi)\dot{\varphi}
        + \alpha_1 (\dot{h}(\bm q, \bm v, \varphi)) \\
        \alpha_2 h_e(\bm q, \bm v, \varphi) +\ ^{W}(\dot{\bm J}^{\{\bm {j+1\}}}_{\bm {flight}}(\bm q)\bm v )_z - \ddot{z}^{\{j+1\}}(\varphi)\dot{\varphi}
        + \alpha_1 (\dot{h}(\bm q, \bm v, \varphi)) \\
        \vdots
    \end{bmatrix},
    \label{eq-qrd-standard}
    \end{split}
\normalsize
\end{equation}
where we add one row for each foot in flight; since $\bm J^{\{\bm j\}}_{\bm {flight}}(\bm q) \in \mathbb{R}^{1 \times n_v}$
is a row vector.

\section{Data-Based Terrain Estimator}

\subsection{Network Architecture}

Section~\ref{subsection:model-terrain} discussed the existing literature on modeling terrain with machine learning.
In this section, we propose a new approach in the same vein. Instead of constructing a sparse graph manually,
we use a dense graph in which every node is connected. It is possible to achieve such a model using Transformer architecture~\cite{Vaswani2017}.
In this sense, each node in the graph becomes a token. The
tokens are constructed by combining the proprioceptively available measurements: joint positions, $\bm q$, joint velocities, $\dot{\bm q}$,
joint torques, $\bm \tau$.
The full attention mechanism is not suited for long sequences. Instead, we use one of the efficient transformers,
Linformer~\cite{Wang2020}. Linformer
reduces the $O(n^2)$ complexity to $O(n)$, allowing us to use it for long sequences with little loss of expression.

The proposed network is illustrated in Figure~\ref{fig:transformer_arch}. Network architecture draws heavy inspiration
from the vision transformer~\cite{dosovitskiy2020vit}.
Measurements of the joint quantities over a time window are concatenated with learned spatial-temporal position embeddings and fed into the network.
Using such a model, terrain estimations can be provided to the ID-CBF-QP described in
Section~\ref{subsection:id-cbf-qp}:
Friction cone constraints, (\ref{eq-qrd-cstr-fcone-x}) and (\ref{eq-qrd-cstr-fcone-y}),
can be improved by updating $\mu$;
CBF ground-clearance constraint, (\ref{eq-qrd-cstr-cbf}), can be improved by updating $z(\varphi)$ in
(\ref{eq-cbf-groundClearance}).

\begin{figure}[thbp]
    \centering
    \includegraphics[width=0.48\textwidth, keepaspectratio]{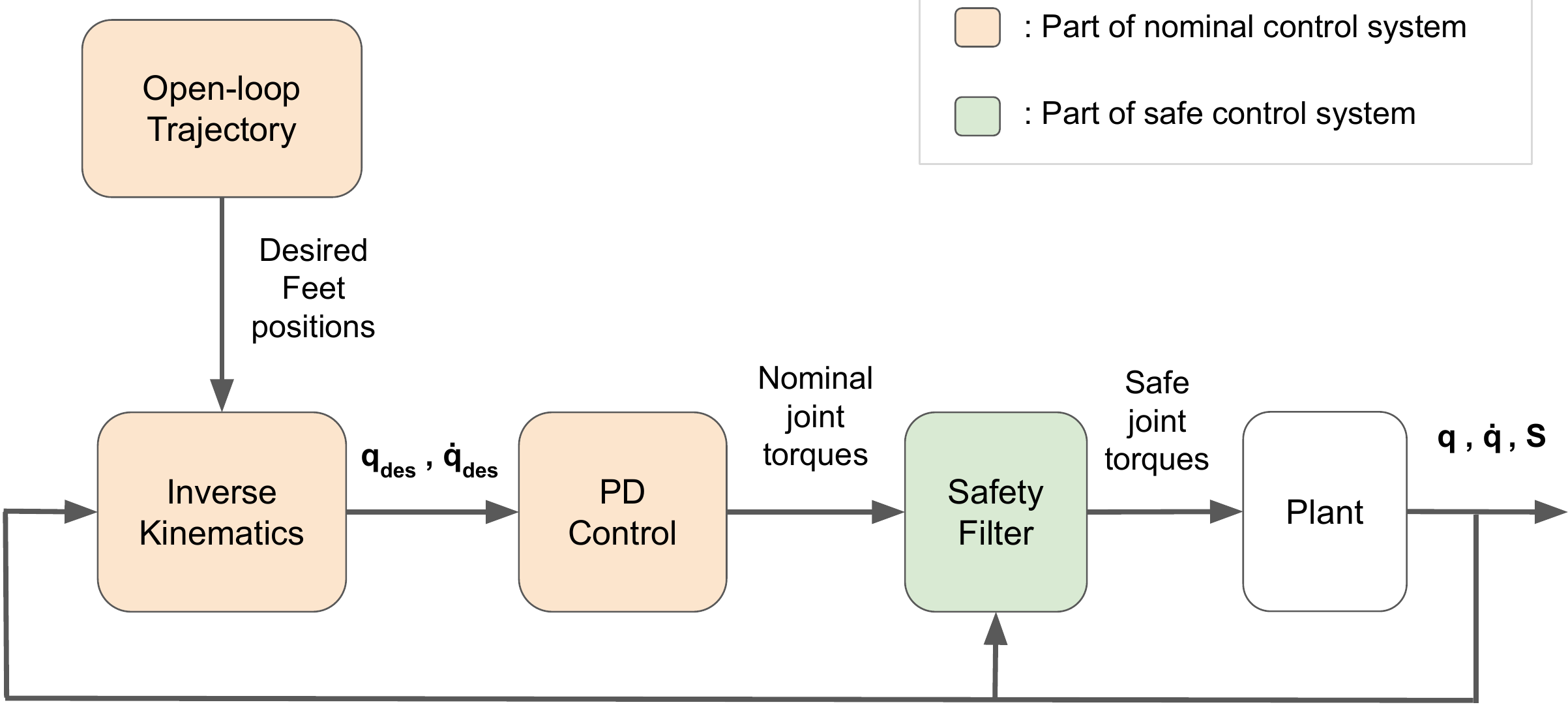}
    \caption{Control diagram of the quadruped robot.}
    \label{fig:diagram-nominal}
\end{figure}

\subsection{Data Collection}
To train the terrain estimation model described in the previous section, large amounts of data are required.
We use Nvidia Omniverse Isaac~\cite{Makoviychuk2021Isaac}, a GPU-accelerated simulator that can achieve an order
of faster simulations than the common CPU-based simulators.

Building on the work of  ~\cite{Rudin2021}, we can use Nvidia Isaac to simulate and gather data for legged robots.
A simulation with 4096 parallel Unitree A1 quadrupeds reaches 100,000 steps
per second on an Nvidia RTX 3080 12GB GPU.
We record the simulation observations to create a dataset.
Table~\ref{table:isaac-collect-subsample} displays the refined input for the model.

\section{Quadruped Implementation}

We have implemented
a software package to work on mechanical systems via optimal control~\cite{source-code}.
Our software package uses PyBullet~\cite{Coumans2021} for simulation and Pinocchio~\cite{Carpentier2019} for rigid body computations.
PyBullet is used only for simulation and observation; all calculations are carried out via Pinocchio.
This enables an easy transition to an actual robot, where highly efficient and embedded-ready Pinocchio is fully utilized.
To solve the ID-CBF-QP, we use OSQP~\cite{osqp}.

% \subsection{Simulation}

The selected quadruped robot is Unitree A1, it has 12 actuated joints, 3 for each leg; it weighs 12 kilograms. The nominal
controller, shown in Figure~\ref{fig:diagram-nominal}, achieves a trotting gait with open-loop feet trajectories.

% \subsection{Moving From Simulation to an Actual Robot}

% The control signal, $\bm u$, can be expressed as an affine function of other variables.
% Though the problem size decreases moderately, the reductions in solution time are significant due to $O(n^3)$ time complexity~\cite{adelprete2016}.
% An embedded-ready, C++ implementation of this approach can be found in TSID~\cite{adelprete2016}.

\section{Results and Discussion}

\subsection{Friction Cone}
The ground reaction forces predicted by the ID-CBF-QP from Section~\ref{subsection:id-cbf-qp} are close to the actual values from the simulation:
the vertical component has a mean absolute error of 3.051 N;
the lateral components have a mean absolute error of 1.847 N.

\begin{figure}[thbp]
  \centering
    \includegraphics[width=0.45\textwidth, keepaspectratio]{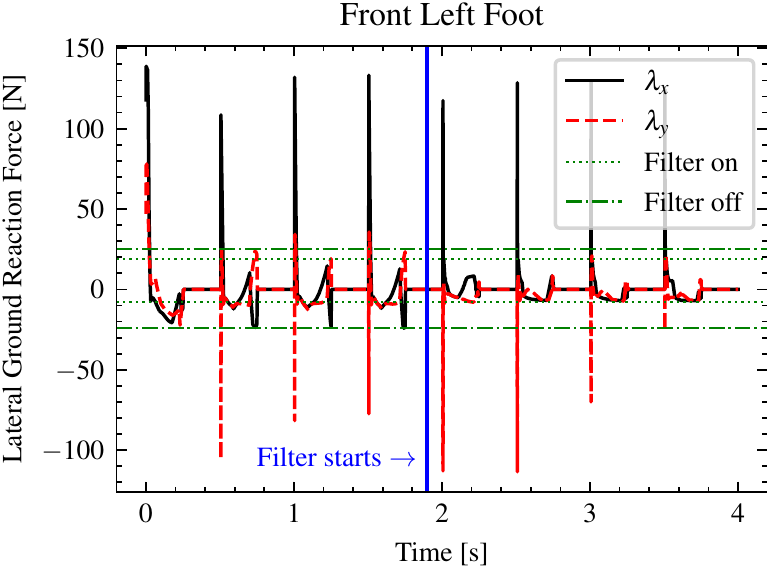}
    \caption{Decrease in lateral forces due to friction constraint after the filter is activated.
    The robot walks normally until the filter starts; then filter reduces lateral (tangential) forces.
    We observe spikes as the foot impacts the ground.
    }
  \label{fig:grf-lateral-diff}
\end{figure}

To demonstrate the effectiveness of the friction constraint, a low friction value is set for the constraints:
$\mu = 0.2$. The resulting control and ground reaction forces are shown for an arbitrarily selected foot in Figure~\ref{fig:grf-lateral-diff}.
In the figure, only the lateral forces are shown to keep it comprehensible.
Since our nominal controller uses inverse kinematics, it does not take forces into account.
We compute inverse dynamics and apply friction cone constraints by introducing the safety filter;
thus, allowing us to eliminate foot sliding and enable safer locomotion.

\subsection{Ground Clearance}

\begin{figure}[thbp]
  \centering
    \includegraphics[width=0.45\textwidth, keepaspectratio]{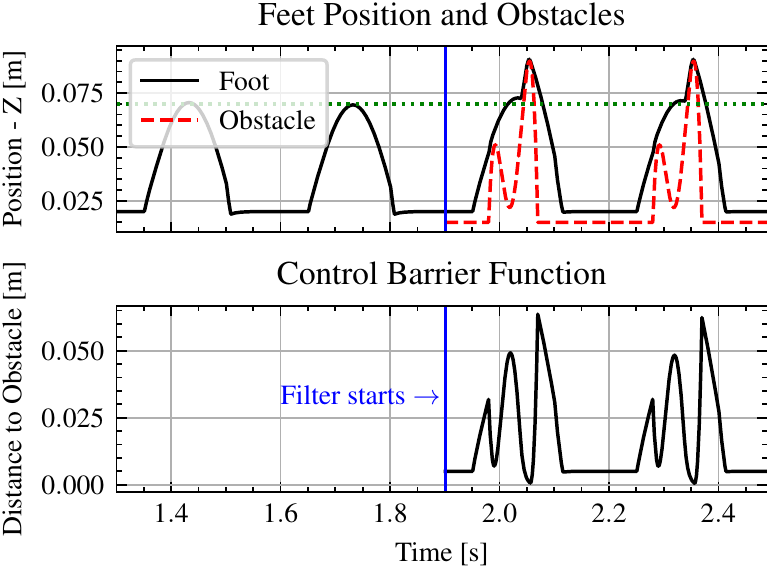}
    \caption{Simulation results with the ground-clearance constraint as a 4\textsuperscript{th} order polynomial.
    The ECBF constraint is active after the filter starts, and it is only defined during the flight phases of the foot.}
  \label{fig:qrd-ecbf-2peaks}
\end{figure}

Figure~\ref{fig:qrd-ecbf-2peaks} shows the simulation results in the presence of the ground-clearance constraint from Section~\ref{section:ECBF}.
In the figure, we observe that the ECBF framework
works as expected and keeps the foot clear of the obstacle with minimal interference to the nominal control signal.
We have used a polynomial for demonstration. However, this is not necessary because our formulation only requires the obstacle profile to be
twice differentiable. ID-CBF-QP can become infeasible for aggressive obstacle profiles due to limited joint torques.

\subsection{Terrain Estimator}

The model can predict the friction coefficient
with a mean absolute error of 0.0720 using randomly sampled 3264 data points from the test split.
Figure~\ref{fig:tt-pred-reg} shows the regression performance of the predictions.
In our work, we obtained the good results using data from a range of 0.6-2 seconds.
In such time spans, there are hundreds of tokens. This justifies using efficient transformers
to model them as long sequences.

\section{Conclusion and Outlook}

This study presents a novel approach to safety-critical control of legged robots.
We have developed a flexible optimal control software package for legged robots.
We have included several safety criteria by defining constraints for the optimization problem. They cover some of the major
sources of locomotion failures. In addition, we have developed a machine-learning model to estimate
terrain properties using proprioceptive sensors. The developed terrain estimator can be added to most robots to increase their contact awareness which
can then be used for high-performance control.

Our work is limited to simulation.
Experimental work would be beneficial in the future. Thus, intentional or unintentional simplifications from the simulation could be
verified and fixed. The terrain estimator could be extended to height map estimation as a future work.
Also, there is no established method to evaluate the performance of legged robots; therefore
a locomotion benchmark would accelerate progress in the field. Furthermore, improved definitions of locomotion safety,
especially ones with a probabilistic sense would be valuable.

\addtolength{\textheight}{-0.25cm}   % This command serves to balance the column lengths

%%%%%%%%%%%%%%%%%%%%%%%%%%%%%%%%%%%%%%%%%%%%%%%%%%%%%%%%%%%%%%%%%%%%%%%%%%%%%%%%

%%%%%%%%%%%%%%%%%%%%%%%%%%%%%%%%%%%%%%%%%%%%%%%%%%%%%%%%%%%%%%%%%%%%%%%%%%%%%%%%

%%%%%%%%%%%%%%%%%%%%%%%%%%%%%%%%%%%%%%%%%%%%%%%%%%%%%%%%%%%%%%%%%%%%%%%%%%%%%%%%

% \bibliographystyle{IEEEtran}
% \bibliography{IEEEabrv,mybibfile}

\begin{thebibliography}{10}
    \providecommand{\url}[1]{#1}
    \csname url@samestyle\endcsname
    \providecommand{\newblock}{\relax}
    \providecommand{\bibinfo}[2]{#2}
    \providecommand{\BIBentrySTDinterwordspacing}{\spaceskip=0pt\relax}
\providecommand{\BIBentryALTinterwordstretchfactor}{4}
\providecommand{\BIBentryALTinterwordspacing}{\spaceskip=\fontdimen2\font plus
\BIBentryALTinterwordstretchfactor\fontdimen3\font minus
\fontdimen4\font\relax}
\providecommand{\BIBforeignlanguage}[2]{{%
\expandafter\ifx\csname l@#1\endcsname\relax
\typeout{** WARNING: IEEEtran.bst: No hyphenation pattern has been}%
\typeout{** loaded for the language `#1'. Using the pattern for}%
\typeout{** the default language instead.}%
\else
\language=\csname l@#1\endcsname
\fi
#2}}
\providecommand{\BIBdecl}{\relax}
\BIBdecl

\bibitem{Grizzle2021Models}
Y.~Gong and J.~W. Grizzle, ``Zero dynamics, pendulum models, and angular
momentum in feedback control of bipedal locomotion,'' \emph{Journal of
Dynamic Systems, Measurement, and Control}, vol. 144, no.~12, p. 121006,
2022.

\bibitem{Zaytsev2018}
P.~Zaytsev, W.~Wolfslag, and A.~Ruina, ``The boundaries of walking stability:
Viability and controllability of simple models,'' \emph{IEEE Transactions on
Robotics}, vol.~34, no.~2, pp. 336--352, 2018.

\bibitem{Carpentier2018}
J.~Carpentier and N.~Mansard, ``Multicontact locomotion of legged robots,''
\emph{IEEE Transactions on Robotics}, vol.~34, no.~6, pp. 1441--1460, 2018.

\bibitem{Siekmann2021BlindStair}
J.~Siekmann, K.~Green, J.~Warila, A.~Fern, and J.~Hurst, ``Blind bipedal stair
traversal via sim-to-real reinforcement learning,'' \emph{arXiv preprint
  arXiv:2105.08328}, 2021.

  \bibitem{Miki2022perfect-rl}
  T.~Miki, J.~Lee, J.~Hwangbo, L.~Wellhausen, V.~Koltun, and M.~Hutter,
  ``Learning robust perceptive locomotion for quadrupedal robots in the wild,''
  \emph{Science Robotics}, vol.~7, no.~62, p. eabk2822, 2022.

  \bibitem{Ames2019}
  A.~D. Ames, S.~Coogan, M.~Egerstedt, G.~Notomista, K.~Sreenath, and P.~Tabuada,
  ``Control barrier functions: Theory and applications,'' in \emph{2019 18th
  European control conference (ECC)}.\hskip 1em plus 0.5em minus 0.4em\relax
  IEEE, 2019, pp. 3420--3431.

  \bibitem{source-code}
  \BIBentryALTinterwordspacing
  B.~Tosun, ``taslel: Terrain-aware safe legged locomotion,'' 2023. [Online].
  Available: \url{https://github.com/Berk-Tosun/taslel}
  \BIBentrySTDinterwordspacing

  \bibitem{underactuated}
  \BIBentryALTinterwordspacing
  R.~Tedrake, \emph{Underactuated Robotics}, 2023. [Online]. Available:
  \url{https://underactuated.csail.mit.edu}
  \BIBentrySTDinterwordspacing

  \bibitem{Featherstone2008Dynamics}
  R.~Featherstone and D.~E. Orin, \emph{Dynamics}.\hskip 1em plus 0.5em minus
  0.4em\relax Berlin, Heidelberg: Springer Berlin Heidelberg, 2008, pp. 35--65.

  \begin{figure}[thbp]
    \centering
      \includegraphics[width=0.4\textwidth]{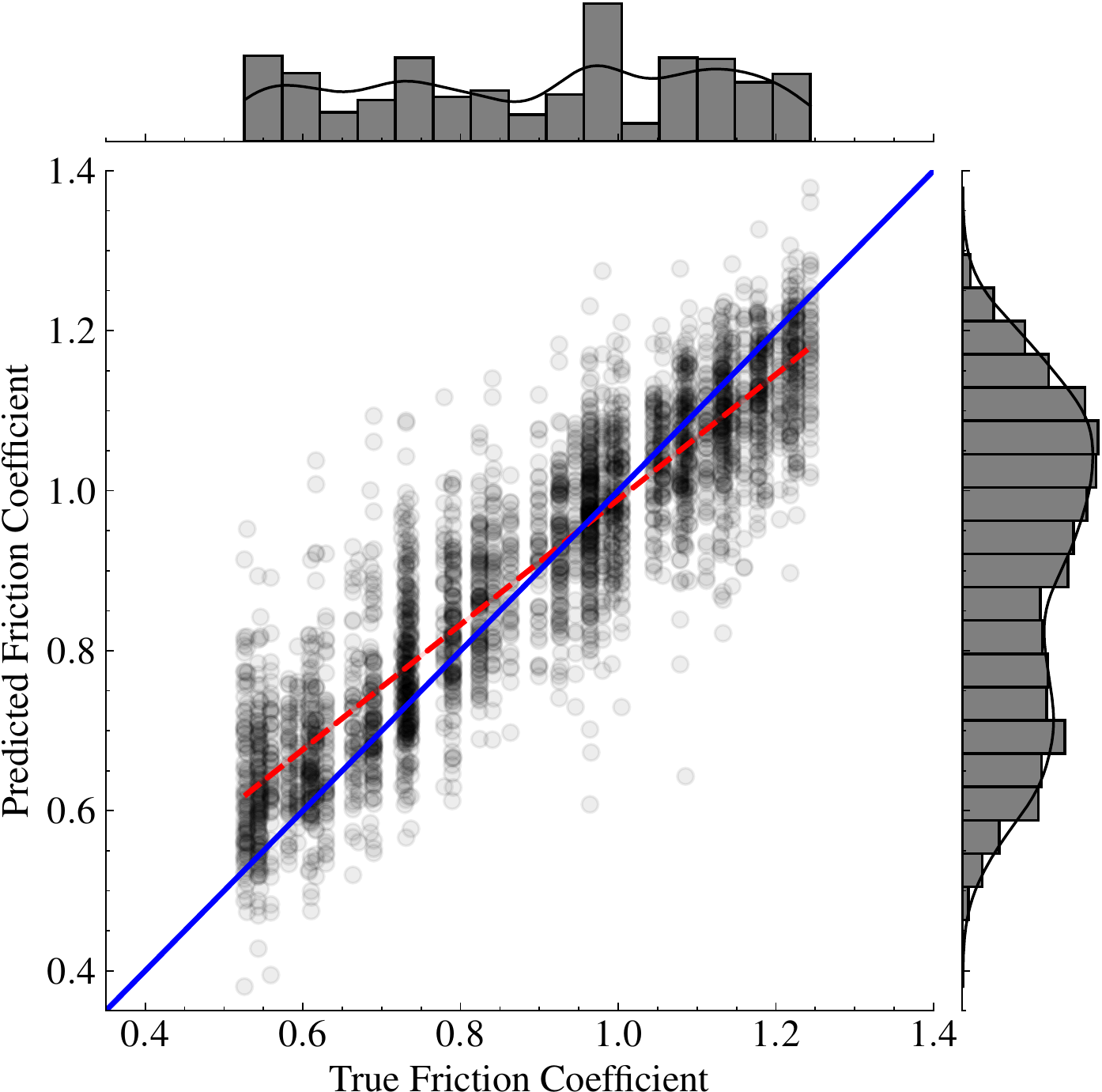}
      \caption{Regression performance of the terrain estimator. Error residuals
      are symmetrically distributed, implying a good fit.}
    \label{fig:tt-pred-reg}
  \end{figure}

\bibitem{Wieber2006dynamics}
P.~B. Wieber, ``Holonomy and nonholonomy in the dynamics of articulated
  motion,'' \emph{Fast Motions in Biomechanics and Robotics: Optimization and
  Feedback Control}, pp. 411--425, 2006.

\bibitem{Wieber2008}
P.-B. Wieber, ``Viability and predictive control for safe locomotion,'' in
  \emph{2008 IEEE/RSJ International Conference on Intelligent Robots and
  Systems}.\hskip 1em plus 0.5em minus 0.4em\relax IEEE, 2008, pp. 1103--1108.

\bibitem{Levine2021Lessons}
J.~Ibarz, J.~Tan, C.~Finn, M.~Kalakrishnan, P.~Pastor, and S.~Levine, ``How to
  train your robot with deep reinforcement learning: lessons we have learned,''
  \emph{The International Journal of Robotics Research}, vol.~40, no. 4-5, pp.
  698--721, 2021.

\bibitem{Vukobratovi2004zmp}
M.~Vukobratovi{\'c} and B.~Borovac, ``Zero-moment point—thirty five years of
  its life,'' \emph{International journal of humanoid robotics}, vol.~1,
  no.~01, pp. 157--173, 2004.

\bibitem{Hirukawa2006CWC}
H.~Hirukawa, S.~Hattori, K.~Harada, S.~Kajita, K.~Kaneko, F.~Kanehiro,
  K.~Fujiwara, and M.~Morisawa, ``A universal stability criterion of the foot
  contact of legged robots-adios zmp,'' in \emph{Proceedings 2006 IEEE
  International Conference on Robotics and Automation, 2006. ICRA 2006.}\hskip
  1em plus 0.5em minus 0.4em\relax IEEE, 2006, pp. 1976--1983.

\bibitem{Yan2018}
S.~Yan, Y.~Xiong, and D.~Lin, ``Spatial temporal graph convolutional networks
  for skeleton-based action recognition,'' in \emph{Proceedings of the AAAI
  conference on artificial intelligence}, vol.~32, no.~1, 2018.

\bibitem{Vaswani2017}
A.~Vaswani, N.~Shazeer, N.~Parmar, J.~Uszkoreit, L.~Jones, A.~N. Gomez,
  {\L}.~Kaiser, and I.~Polosukhin, ``Attention is all you need,''
  \emph{Advances in neural information processing systems}, vol.~30, 2017.

\bibitem{Tay2020}
Y.~Tay, M.~Dehghani, S.~Abnar, Y.~Shen, D.~Bahri, P.~Pham, J.~Rao, L.~Yang,
  S.~Ruder, and D.~Metzler, ``Long range arena: A benchmark for efficient
  transformers,'' \emph{arXiv preprint arXiv:2011.04006}, 2020.

\bibitem{Wang2020}
S.~Wang, B.~Z. Li, M.~Khabsa, H.~Fang, and H.~Ma, ``Linformer: Self-attention
  with linear complexity,'' \emph{arXiv preprint arXiv:2006.04768}, 2020.

\bibitem{Lee2020}
J.~Lee, J.~Hwangbo, L.~Wellhausen, V.~Koltun, and M.~Hutter, ``Learning
  quadrupedal locomotion over challenging terrain,'' \emph{Science robotics},
  vol.~5, no.~47, p. eabc5986, 2020.

\bibitem{Reher2020}
J.~Reher, C.~Kann, and A.~D. Ames, ``An inverse dynamics approach to control
  lyapunov functions,'' in \emph{2020 American Control Conference (ACC)}.\hskip
  1em plus 0.5em minus 0.4em\relax IEEE, 2020, pp. 2444--2451.

\bibitem{Featherstone2004}
R.~Featherstone, ``An empirical study of the joint space inertia matrix,''
  \emph{The International Journal of Robotics Research}, vol.~23, no.~9, pp.
  859--871, 2004.

\bibitem{Grandia2021}
R.~Grandia, A.~J. Taylor, A.~D. Ames, and M.~Hutter, ``Multi-layered safety for
  legged robots via control barrier functions and model predictive control,''
  in \emph{2021 IEEE International Conference on Robotics and Automation
  (ICRA)}.\hskip 1em plus 0.5em minus 0.4em\relax IEEE, 2021, pp. 8352--8358.

\bibitem{Nguyen2016stepping}
Q.~Nguyen, A.~Hereid, J.~W. Grizzle, A.~D. Ames, and K.~Sreenath, ``3d dynamic
  walking on stepping stones with control barrier functions,'' in \emph{2016
  IEEE 55th Conference on Decision and Control (CDC)}.\hskip 1em plus 0.5em
  minus 0.4em\relax IEEE, 2016, pp. 827--834.

\bibitem{dosovitskiy2020vit}
A.~Dosovitskiy, L.~Beyer, A.~Kolesnikov, D.~Weissenborn, X.~Zhai,
  T.~Unterthiner, M.~Dehghani, M.~Minderer, G.~Heigold, S.~Gelly \emph{et~al.},
  ``An image is worth 16x16 words: Transformers for image recognition at
  scale,'' \emph{arXiv preprint arXiv:2010.11929}, 2020.

\bibitem{Makoviychuk2021Isaac}
V.~Makoviychuk, L.~Wawrzyniak, Y.~Guo, M.~Lu, K.~Storey, M.~Macklin,
  D.~Hoeller, N.~Rudin, A.~Allshire, A.~Handa \emph{et~al.}, ``Isaac gym: High
  performance gpu-based physics simulation for robot learning,'' \emph{arXiv
  preprint arXiv:2108.10470}, 2021.

\bibitem{Rudin2021}
N.~Rudin, D.~Hoeller, P.~Reist, and M.~Hutter, ``Learning to walk in minutes
  using massively parallel deep reinforcement learning,'' in \emph{Conference
  on Robot Learning}.\hskip 1em plus 0.5em minus 0.4em\relax PMLR, 2022, pp.
  91--100.

\bibitem{Coumans2021}
E.~Coumans and Y.~Bai, ``Pybullet, a python module for physics simulation for
  games, robotics and machine learning,'' 2016.

\bibitem{Carpentier2019}
J.~Carpentier, G.~Saurel, G.~Buondonno, J.~Mirabel, F.~Lamiraux, O.~Stasse, and
  N.~Mansard, ``The pinocchio c++ library: A fast and flexible implementation
  of rigid body dynamics algorithms and their analytical derivatives,'' in
  \emph{2019 IEEE/SICE International Symposium on System Integration
  (SII)}.\hskip 1em plus 0.5em minus 0.4em\relax IEEE, 2019, pp. 614--619.

\bibitem{osqp}
B.~Stellato, G.~Banjac, P.~Goulart, A.~Bemporad, and S.~Boyd, ``Osqp: An
  operator splitting solver for quadratic programs,'' \emph{Mathematical
  Programming Computation}, vol.~12, no.~4, pp. 637--672, 2020.

\end{thebibliography}


\end{document}